\documentclass[lettersize,journal]{IEEEtran}
\usepackage{amsmath,amsfonts}
\usepackage{algorithmic}
\usepackage{algorithm}
\usepackage{array}
\usepackage[caption=false,font=normalsize,labelfont=sf,textfont=sf]{subfig}
\usepackage{textcomp}
\usepackage{stfloats}
\usepackage{url}
\usepackage{verbatim}
\usepackage{graphicx}
\usepackage{cite}

\newcommand{\etal}{\textit{et al}.~}
\newcommand{\ieno}{\textit{i}.\textit{e}.}
\newcommand{\etcno}{\textit{etc}} 

\usepackage{xcolor}

\newcommand{\tcr}{\textcolor{red}}

\usepackage{hyperref}
\usepackage{enumitem}

\usepackage{amsfonts}
\usepackage{enumitem}
\usepackage{multirow}
\usepackage{booktabs} 
\usepackage{bbding}

\begin{document}

\title{Skeleton-Based Mutually Assisted Interacted Object Localization and Human Action Recognition}

\author{Liang~Xu, Cuiling~Lan,~\IEEEmembership{Member,~IEEE,}
        Wenjun~Zeng,~\IEEEmembership{Fellow,~IEEE,}
        and~Cewu~Lu,~\IEEEmembership{Member,~IEEE}
\IEEEcompsocitemizethanks{\IEEEcompsocthanksitem This work was done while Liang Xu was an intern with Microsoft Research Asia.\protect\\
\IEEEcompsocthanksitem Liang Xu and Cewu Lu are with the Department of Electrical and Computer Engineering, Shanghai Jiao Tong University, Shanghai, 200240, China. E-mail: liangxu@sjtu.edu.cn; lucewu@sjtu.edu.cn\protect\\
\IEEEcompsocthanksitem Cuiling Lan and Wenjun Zeng are with Microsoft Research Asia. E-mail: culan@microsoft.com; zengw2011@hotmail.com\\
\IEEEcompsocthanksitem Corresponding author: Cuiling Lan.}}

\markboth{IEEE TRANSACTIONS ON MULTIMEDIA, 2022}%
{Shell \MakeLowercase{\textit{et al.}}: A Sample Article Using IEEEtran.cls for IEEE Journals}

\IEEEpubid{\begin{minipage}{\textwidth}\ \\[12pt] \centering
  1920-9210 \copyright 2020 IEEE. Personal use is permitted, but republication/redistribution requires IEEE permission.\\
  Seehttps://www.ieee.org/publications/rights/index.html for more information.
\end{minipage}} 

\maketitle

\begin{abstract}
Skeleton data carries valuable motion information and is widely explored in human action recognition. 
However, not only the motion information but also the interaction with the environment provides discriminative cues to recognize the action of persons.
In this paper, we propose a joint learning framework for mutually assisted ``interacted object localization'' and ``human action recognition'' based on skeleton data. 
The two tasks are serialized together and collaborate to promote each other, where preliminary action type derived from skeleton alone helps improve interacted object localization, which in turn provides valuable cues for the final human action recognition.
Besides, we explore the temporal consistency of interacted object as constraint to better localize the interacted object with the absence of ground-truth labels.
Extensive experiments on the datasets of SYSU-3D, NTU60 RGB+D, Northwestern-UCLA and UAV-Human show that our method achieves the best or competitive performance with the state-of-the-art methods for human action recognition.
Visualization results show that our method can also provide reasonable interacted object localization results.
\end{abstract}


\section{Introduction}

Human action recognition is a pivotal problem in video-based tasks, such as human-computer interaction, intelligent surveillance, video understanding and robotics~\cite{poppe2010survey,weinland2011survey,aggarwalsurvey}.
In particular, skeleton-based action recognition shows its superiority and develops rapidly in recent years~\cite{st-gcn,two-stream,li2019actional,semantics2019,zhang2020context,cheng2020skeleton,liu2020disentangling,su2020predict,chengdecoupling,zhang2020deep,zhu2019cuboid,hu2019joint,avola20192,liu2020multi,yang2020hierarchical}. Skeleton data contains the 2D or 3D coordinates of the human key joints, which is a high level, efficient, robust and privacy-friendly representation for human poses and motion information. Besides, the access of skeleton data becomes easier thanks to the development of depth cameras and human pose estimation algorithms.

\begin{figure}[t]
\begin{center}
  \includegraphics[width=1.0\linewidth]{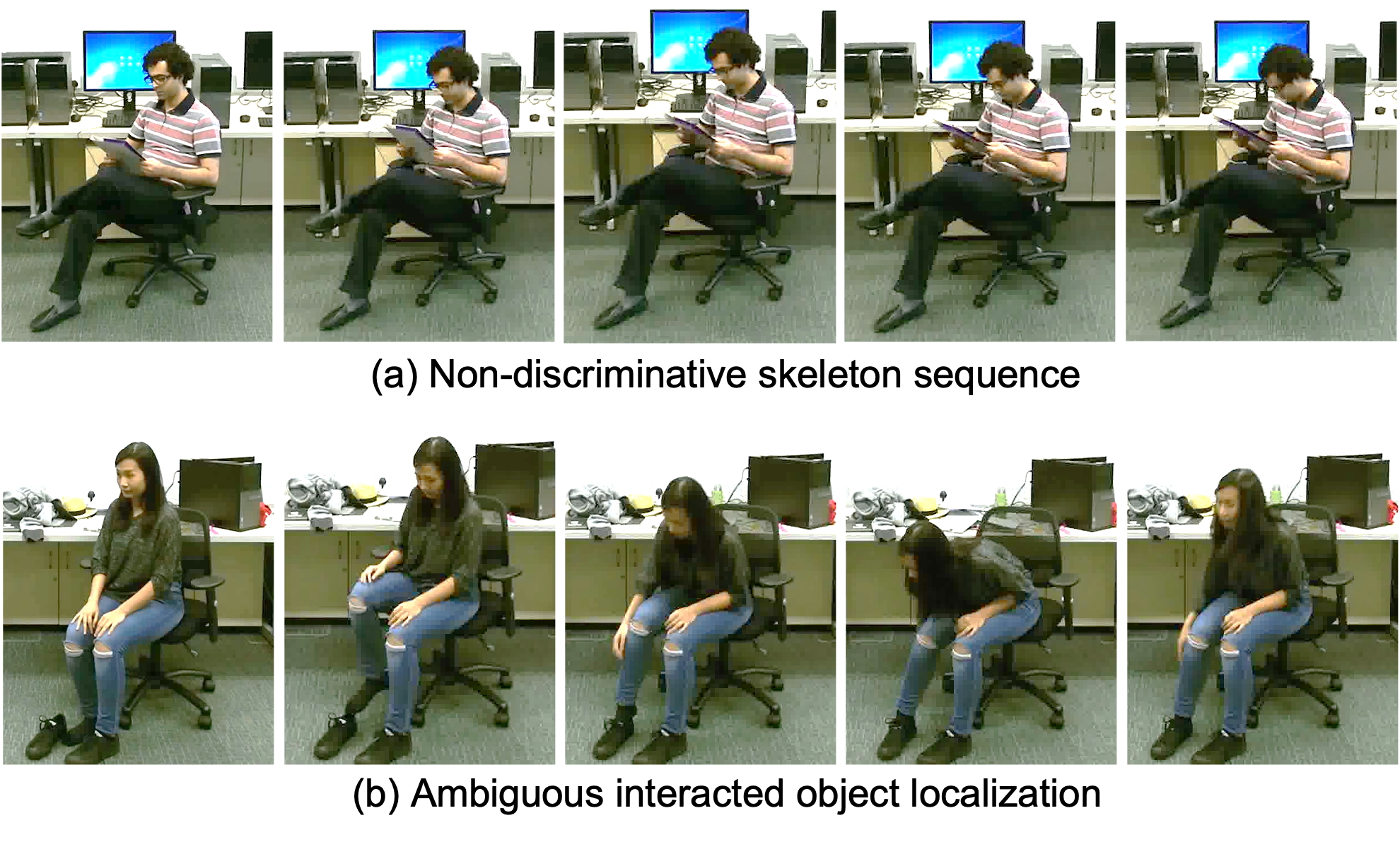}
\end{center}
  \caption{Two examples of action ``read'' and ``wear a shoe''. For (a), the skeleton sequence alone is not discriminative enough for recognizing ``read''. Integrating object information, \ieno, the book, can facilitate the recognition of ``read''. For (b), the skeleton data could be ambiguous for localizing the interacted object without the knowledge of the action type (both the shoe and the chair could be interacted with).}
\label{fig:firstpage}
\end{figure}

Powered by deep learning especially graph convolutional network, skeleton data has been proved to be effective for human action recognition. However, skeleton input alone may encounter several problems.
First, skeleton sequences could be deficient and ambiguous under certain circumstances. For example, given a standing-still person talking to another person or watching a television, their action categories differ greatly yet the skeleton sequences have little difference.
Second, these methods neglect the human-environment interactions, which play indispensable roles to model human actions~\cite{something-else,tang2020asynchronous}. Human-object interactions account for the majority of daily human activities~\cite{pastanet} and the object knowledge can serve as strong and complementary cues.
Third, skeleton data cannot handle those fine-grained action recognition tasks, such as analyzing the behavior of customers picking up commodities in a store.
Therefore, aggregating skeleton data with its interacted object is intuitively helpful in dealing with the aforementioned challenges.

Nonetheless, interacted object localization in videos is still an open and less explored problem. 
For \emph{image-based} human object interaction detection, Gkioxari \etal~\cite{gkioxari2018detecting} predict the interacted object location based on the appearance feature of a detected person with the supervision of volumes of labeled interacted person-object box pairs.
\IEEEpubidadjcol
However, in video action datasets, labeling the bounding boxes of interacted person-object pairs is quite expensive and frustrating.
Luckily, skeleton sequence provides some cues for localizing where the interacted object is, such as the distance between the human skeleton and the objects, the active human parts for localizing a rough object area.
Besides, temporal motion and consistency of interacted object could also be explored to benefit the localization.

In this paper, in order to jointly exploit the interacted object and skeleton data for robust human action recognition, we introduce an auxiliary task called ``\textit{skeleton-based interacted object localization}'', \ieno, localize the interacted object of a given person based on the skeleton data in an unsupervised manner.
Skeleton-based interacted object localization shares the same difficulties as human action recognition. As demonstrated in Fig.~\ref{fig:firstpage}(b), the girl is sitting on a chair and wearing a shoe simultaneously. The skeleton data may be ambiguous and insufficient for localizing the interacted object, unless the action type is given. Moreover, action type carries valuable semantic information to indicate the interaction style and can help filter improbable objects, thus could boost the object localization performance.

We design a joint learning framework for ``interacted object localization" and ``human action recognition". The two tasks are serialized together and collaborate to mutually assist each other. Fig.~\ref{fig:pipeline} illustrates our framework with skeleton sequences and object proposals as input. An object detector is first applied on each frame to obtain the \textit{object proposals/candidates} (\ieno, positions and categories of objects) for the subsequent interacted object localization. A skeleton-only action recognition model generates a preliminary action classification score, followed by a joint learning procedure of (a) Action Type Assisted Interacted Object Localization and (b) Interacted Object Assisted Human Action Recognition.

We summarize our main contributions as follows.
\begin{itemize}[leftmargin=*,noitemsep,nolistsep]
    \item We design a joint learning framework for interacted object localization and human action recognition based on skeleton data without ground-truth of interacted object.
    \item We explore the preliminary action type cues and the consistency characteristic of interacted object across time to promote the reliable localization of interacted object, which in turn helps action recognition. 
\end{itemize}

We evaluate the performance of our proposed method on three widely used datasets for skeleton-based action recognition: SYSU-3D~\cite{sysu_dataset}, NTU60 RGB+D~\cite{ntudataset}, Northwestern-UCLA~\cite{ucla_dataset} and UAV-Human~\cite{uav_human}. Extensive experiments show that: 
(1) integrating interacted object information can significantly boost the performance of human action recognition;
(2) incorporating action type cues with skeleton data enhances the performance of interacted object localization, which in turn helps human action recognition;
(3) reasonable interacted object localization results are obtained even without interacted object labels.
We will release our code and models upon acceptance.

\section{Related Work}
\subsection{Video Action Understanding}
Action recognition is fundamental in video-based tasks with many approaches proposed~\cite{ji20123d,tran2014c3d,taylor2010convolutional,varol2017long,carreira2017quo,qiu2017learning,diba2018spatio,tran2018closer,xie2017rethinking,christoph2016spatiotemporal} and  datasets~\cite{soomro2012ucf101,kay2017kinetics,ntudataset,sysu_dataset,ucla_dataset,gu2018ava,caba2015activitynet,gupta2020quo}. We notice that there is also a trend for more fine-grained action understanding, from video classification~\cite{ji20123d,tran2014c3d} to spatial-temporal action detection~\cite{gu2018ava,feichtenhofer2019slowfast,hou2017tube,tang2020asynchronous}, and human-part level action recognition~\cite{pastanet}. However, so far, interacted object localization together with skeleton motion for robust human action recognition is less explored. We propose a joint learning framework of interacted object localization and action recognition and enable their mutual promotion.

\subsection{Skeleton-based Action Recognition}
Skeleton-based action recognition attracts much attention. Traditional approaches~\cite{hussein2013human,wang2012mining,vemulapalli2014human} design hand-crafted features for action classification.
Hussein \etal~\cite{hussein2013human} use the covariance matrices of joint trajectories as a discriminative descriptor.
Wang \etal~\cite{wang2012mining} propose an actionlet ensemble model to represent each action and then capture the intra-class variance.
Vemulapalli \etal~\cite{vemulapalli2014human} propose a skeletal representation to explicitly model the 3D geometric relationships between different body parts using rotations and translations in 3D space.


\begin{figure*}
\begin{center}
  \includegraphics[width=1.0\linewidth]{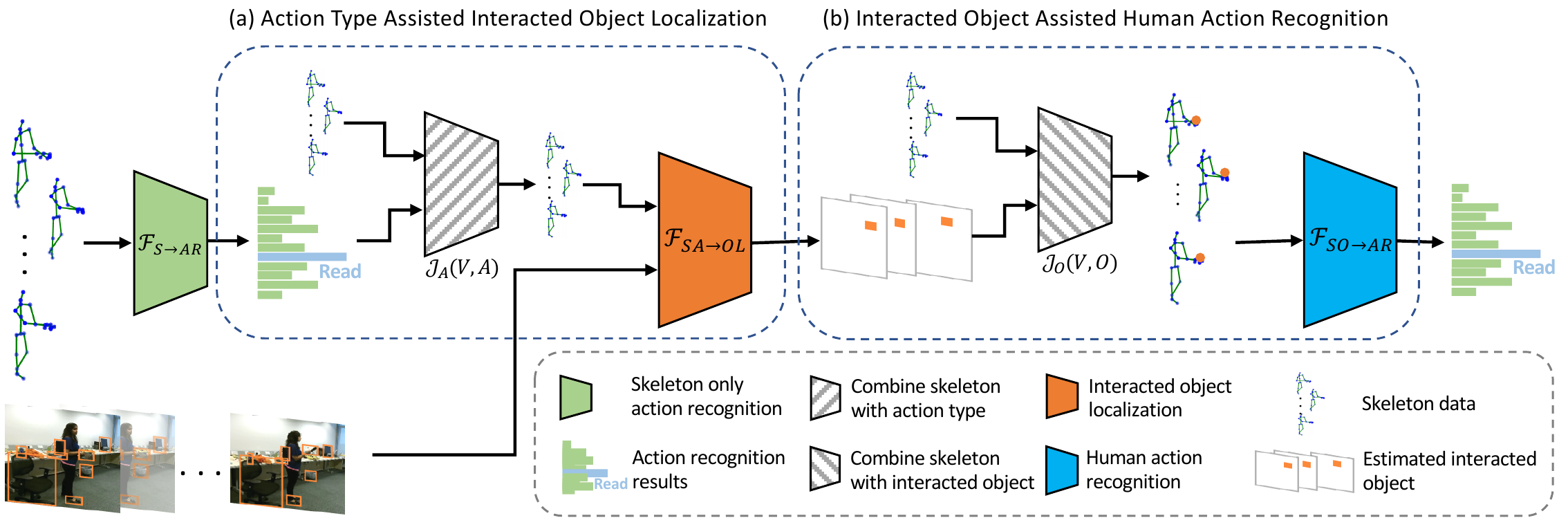}
\end{center}
  \caption{Overview of our proposed joint task learning framework for robust human action recognition. Given a video sequence, we have its skeleton sequence and frame-level object detection results (obtained from an off-the-shelf object detector). A $\mathcal{F}_{S\to AR}$ model, \ieno, skeleton-only human action recognition model, generates a preliminary action classification score, followed by a joint learning procedure of (a) Action Type Assisted Interacted Object Localization and (b) Interacted Object Assisted Human Action Recognition. The two tasks are serialized together and collaborate to mutually promote each other.}
\label{fig:pipeline}
\end{figure*}

With the development of CNNs and RNNs, deep learning based methods significantly outperform the traditional methods. CNN-based methods~\cite{du2015hierarchical,ntudataset,liu2016spatio,song2016end,viewadaptive,li2018independently,cao2018skeleton} map a skeleton sequence into an image-like representation and use the network such as ResNet to extract the spatial and temporal features of the sequence for action recognition. RNN-based methods~\cite{liu2017two,kim2017interpretable,ke2017new,liu2017enhanced,li2017skeleton} take each frame as the input of a time step in recurrent neural networks to model the spatial and temporal correlations for action recognition.

Thanks to the development of graph convolutional networks to process structured data, GCN-based methods~\cite{st-gcn,liu2018recognizing,two-stream,li2019actional,shi2019skeleton,semantics2019,zhang2020context,cheng2020skeleton,liu2020disentangling,su2020predict,chengdecoupling,chen2021channel} have been demonstrated to be effective for skeleton-based action recognition. The latest GCN-based methods can be divided into three categories.
First, some methods design sophisticated GCN structures to better model the spatial and temporal features~\cite{liu2020disentangling,zhang2020context,chen2021channel}.
Liu \etal~\cite{liu2020disentangling} leverage dense cross space-time edges as skip connections for unbiased long-range and unobstructed information propagation in spatial-temporal dimension.
Zhang \etal~\cite{zhang2020context} propose a context-aware GCN to consider a context term for each vertex by integrating information of all other vertices.
Chen \etal~\cite{chen2021channel} propose a channel-wise topology refinement graph convolution to learn different topologies and aggregate joint features in different channels.
Second, some models aim at reducing computational cost for skeleton data~\cite{semantics2019,cheng2020skeleton,chengdecoupling}.
Zhang \etal~\cite{semantics2019} explicitly introduce the high level semantics of joints into the network and build a strong and lightweight baseline for action recognition.
Cheng \etal~\cite{cheng2020skeleton,chengdecoupling} propose the lightweight graph operations to enhance the graph modeling ability.
Cheng \etal~\cite{chengdecoupling} propose decoupling GCN to enhance the graph modeling ability without extra computation.
Third, other works handle unsupervised learning~\cite{su2020predict}, self-supervised learning~\cite{Lin_2020} of skeleton-based action recognition.

In this work, we attempt to integrate the interacted object into the skeleton-based action recognition framework to strengthen the human action recognition. A framework for joint training of interacted object localization and human action recognition is proposed, where the two tasks collaborate to promote each other.

\subsection{Interacted Object Localization}
In image-based scenarios, Gkioxari \etal~\cite{gkioxari2018detecting} predict the interacted object location based on the appearance feature of a detected person with the supervision of volumes of interacted person-object box pairs. Li \etal~\cite{li2019transferable} learn a binary score of interactivenss as a general knowledge for human object interaction detection. In~\cite{nmp}, they hand-craft a rule to determine the interactiveness according to distance between humans and objects. In video-based scenarios, interacted object localization is barely explored due to the absence of interacted human-object bounding boxes. In this work, we study the interacted object localization for action recognition based on skeleton data, without requiring ground-truth locations of the interacted object.

\subsection{Joint Task Learning}
Generic joint task learning is common and widely studied in computer vision. Many tasks have been studied such as the joint tasks of detection and classification~\cite{girshick2015fast,he2017mask}, depth estimation and image decomposition~\cite{kim2016unified}, segmentation and classification~\cite{misra2016cross}, and semantic segmentation and depth estimation from monocular images~\cite{ladicky2014pulling,wang2015towards,kendall2018multi,zhang2018joint}.
In this paper, we propose a joint task learning framework for interacted object localization and human action recognition based on skeleton data. 

\section{Proposed Joint Learning Framework}
In this section, we formulate the proposed joint learning framework for skeleton-based interacted object localization and human action recognition. Fig.~\ref{fig:pipeline} illustrates the proposed framework, where the \textit{action type assisted interacted object localization} and the \textit{interacted object assisted human action recognition} are serialized and collaborate to mutually promote each other. Given a video clip $X = \{ x_t | t = 1,...,T\}$ with $T$ frames, an object detector is applied on each frame to obtain the object candidates. The human skeleton data is captured by depth cameras. The object candidates are denoted by $O = \{o_{t,i} | t = 1,...,T, i=1,...,N_o\}$ and the human skeletons are denoted by $V = \{v_{t,i} | t = 1,...,T, i = 1,...,N_v\}$, where $N_o$ and $N_v$ indicate the number of objects and human joints, respectively. If one frame contains more than one person, we split the frame to make sure each frame contains only one human skeleton as in~\cite{semantics2019}.
Hereinafter, $S$$\to$$AR$ denotes skeleton-only human action recognition, $SA$$\to$$OL$ denotes integrating skeleton data and action type for interacted object localization, and $SO$$\to$$AR$ indicates integrating skeleton data and interacted object for human action recognition.

Previous skeleton-based methods~\cite{st-gcn,two-stream,semantics2019,zhang2020context,cheng2020skeleton,liu2020disentangling,chengdecoupling} learn a classification model: $\widetilde A = \mathcal{F}_{S\to AR}(V, \phi_{S\to AR})$ based only on skeleton data $V$, $\phi_{S\to AR}$ denotes the parameters and $\widetilde A$ is the estimated action recognition result. As mentioned before, we serialize the $SA$$\to$$OL$ and $SO$$\to$$AR$ tasks with the following formulation:
\begin{align}
   \widetilde A^0 &= \mathcal{F}_{S\to AR}(V, \phi_{S\to AR}), \label{eq:form0}\\
  \widetilde O_{att} &= \mathcal{F}_{SA\to OL}\left(\mathcal{J}_{A}(V, \widetilde A^{0}), \phi_{SA\to OL}\right), \label{eq:form1}\\
  \widetilde A^{1} &= \mathcal{F}_{SO\to AR} \left(\mathcal{J}_{O}(V, \widetilde O_{att}) , \phi_{SO\to AR} \right). \label{eq:form2}
\end{align}
In Eq.~\ref{eq:form0}, model $\mathcal{F}_{S\to AR}$ generates the preliminary action classification result $\widetilde A^0$. In Eq.~\ref{eq:form1} of the next step, $\mathcal{J}_A$ is a function of combining skeleton data $V$ and action classification result $\widetilde A^{0}$, then $\mathcal{F}_{SA\to OL}$ takes the combined feature to boost the interacted object localization results $\widetilde O_{att}$.
Then in Eq.~\ref{eq:form2}, $\mathcal{J}_O$ integrates the skeleton data $V$ and the interacted object localization result $\widetilde O_{att}$ from the previous step. The integrated feature is fed into $\mathcal{F}_{SO\to AR}$ for further human action recognition to obtain $\widetilde A^1$. $\phi$ denotes the parameters of these models.

We will introduce the $\mathcal{F}_{S\to AR}$ model in Sect.~\ref{sec:s_ar}, which can be any skeleton-only action recognition model like ST-GCN~\cite{st-gcn}, 2s-AGCN~\cite{two-stream}, or SGN~\cite{semantics2019}. The details of $\mathcal{J}_A$ and $\mathcal{F}_{SA\to OL}$ will be presented in Sect.~\ref{sec:sa_ol}, and $\mathcal{J}_O$ and $\mathcal{F}_{SO\to AR}$ will be presented in Sect.~\ref{sec:so_ar}.

\section{Technical Details}

\subsection{Data Preparation}
\centerline{\textbf{Input}: \{$X$\} $\longmapsto$ \textbf{Output}: \{$O$, $V$\}}

Given a video clip $X$ with $T$ frames, an object detector is applied on each frame to obtain the object candidates $O$. In this paper, we employ the Cascade RCNN~\cite{cai2018cascade} model trained on Object365~\cite{shao2019objects365} to localize and detect objects. Object365~\cite{shao2019objects365} is a newly proposed large-scale object detection dataset with 365 object classes, covering richer interactive objects than the COCO dataset~\cite{lin2014microsoft}. For each frame, we select the first $N_o$ (fixed) object candidates by detection scores.
For more than $N_o$ objects, we select those with higher object detection scores; for less than $N_o$ detected objects, we pad to $N_o$ with the existing detected objects.
The object features are composed of the bounding box, the confidence score, and the object category. In this paper, the RGB frames are only used for obtaining detected objects and the remaining modules are purely skeleton-based.

Following the previous works~\cite{semantics2019}, the skeleton data $V$ is captured by depth cameras and provided as 3D coordinates.

\subsection{Skeleton-Only Action Recognition}
\label{sec:s_ar}
\centerline{\textbf{Input}: \{$V$\} $\longmapsto$ \textbf{Output}: \{$\widetilde A$\}}

Previous effective skeleton-based action recognition models like ST-GCN~\cite{st-gcn}, 2s-AGCN~\cite{two-stream}, or SGN~\cite{semantics2019} can be adopted. In this paper, we select SGN for its high computational efficiency. Next we recap the architecture of SGN.

SGN explicitly leverages the joint type and frame index as high-level semantics to enhance the feature representations. Given $V = \{v_{t,i} | t = 1,...,T, i = 1,...,N_v\}$ of $T$ frames and $N_v$ joints for each person, SGN obtains the dynamics representation $z_{t,i}$ of each joint:
\begin{align}
    z_{t,i} = \Phi_1(v_{t,i}) + \Phi_2(v_{t,i}-v_{t-1,i}),
    \label{eq:sar1}
\end{align}
where $v_{t,i}-v_{t-1,i}$ calculates the velocity of temporal neighboring joints, and $\Phi_1$ and $\Phi_2$ are two stacked fully connected (FC) layers to embed the position and velocity of joints to the same high dimensional space.

For the joint-level module, the joint dynamics feature $j_m$ (\ieno, one-hot vectors to indicate the index of the joint $m$) is concatenated with the semantics of joint type $m$ at frame $t$ as $z_{t,m}=z_{t,m}\oplus j_m$. Then, three adaptive GCN layers are stacked to exploit the correlations of different joints. In each GCN layer, the adjacency matrix $S_t$ is calculated by the inner product of the transformed joint features as:
\begin{align}
    S_t(i,j) = \theta_z(z_{t,i})^\top \cdot \phi_z(z_{t,j}),
    \label{eq:sar2}
\end{align}
where $\theta_z(\cdot)$ and $\phi_z(\cdot)$ are both implemented by a FC layer.

For the frame-level module, the frame index semantics (\ieno, one-hot vectors to indicate the index of the frames) are incorporated into the joint representations $z_{t,i}$ to encode the order of the frames. Then the correlations across frames are modeled by two CNN layers. The first CNN layer is acting as a temporal convolution to model the dependencies of frames and the second layer maps the feature embedding to a high dimensional space. After the spatial-temporal feature embedding, a FC layer with Softmax acts as the action classification function.

\subsection{Action Type Assisted Object Localization}
\label{sec:sa_ol}
\centerline{\textbf{Input}: \{($V, \widetilde A$), $O$\} $\longmapsto$ \textbf{Output}: \{$\widetilde O_{att}$\}}

We perform the interacted object localization task as a selection operation from the detected object candidates $O$ based on the skeleton data $V$ and the preliminary action classification result $\widetilde A$ in an unsupervised manner. The action type $\widetilde A$ can serve as the \textit{state}/\textit{attribute} of joints and enhance the representation ability over only 3D coordinates $\langle x, y, z\rangle$. For each joint of a frame, we incorporate the embedded feature of the valuable action type cues together with the feature of the joints as:
\begin{align}
    f_{sa} = \mathcal{J}_A(v_{t,i}, \widetilde{{A}}) = \Phi_j(v_{t,i}) \oplus \Phi_a(\widetilde{{A}}),
    \label{eq:saol1}
\end{align}
where $\widetilde {{A}}$ denotes the action type vector, $v_{t,i}$ denotes the 3D coordinate of the $i^{th}$ joint in the $t^{th}$ frame of the skeleton sequence, $\Phi_j$ and $\Phi_a$ are constructed by two FCs to embed each joint and the action type to a 128-d feature vector, respectively, $\oplus$ denotes the concatenation operation and $f_{sa}$ denotes the combined 256-d feature. 

The interacted object localization task is to select the most likely object from the candidates set $O$. We model the affinity between each joint and each object by calculating the inner-product of their embedded features, similar to Eq.~\ref{eq:sar2}. Therefore, for a person of $N_v$ joints and $N_o$ object candidates, we can obtain the affinity matrix $\mathcal{A} \in \mathbb{R}^{{N_v}\times{N_o}}$. Then, we perform max-pooling across the joints (rows) to obtain the affinity vector $o_{att} \in \mathbb{R}^{N_o}$, which denotes the affinity between the whole person and each object candidate. The Sigmoid operation is applied to normalize the affinity $o_{att}$ to obtain the attention vector over objects as $\widetilde o_{att} \in \mathbb{R}^{N_o}$. We denote the attention for all the $T$ frames as $\widetilde O_{att} \in \mathbb{R}^{T \times {{N_o}}}$. 


\noindent{\bf Temporal Consistency Loss.} We assume that given a video sequence of a person performing one action, the interacted object category is in general consistent over the time. 
Thus we propose a consistency loss across frames, which is defined as the variance of the categories of selected interacted object. 
Suppose the total number of object categories is $N$ (For COCO-80~\cite{lin2014microsoft} dataset, $N=80$ and each video clip has $T$ frames. We first obtain a matrix $A$ of size $T\times N$ based on $\widetilde O_{att}$, where $A(t,i)$ denotes the probability of being category $i$ for frame $t$ as the interacted object. Then, we first calculate the variance along the $T$ dimension for each object category and then obtain the average value for the $N$ object categories as the consistency loss.


\subsection{Interacted Object Assisted Action Recognition}
\label{sec:so_ar}
\centerline{\textbf{Input}: \{($V, \widetilde O_{att}$), $O$\} $\longmapsto$ \textbf{Output}: \{$\widetilde A$\}}

$\widetilde O_{att}$ serves as an attention to fuse the object candidates features. For each frame, we use the attention vector $\widetilde o_{att}$ as weights to fuse the features of the $N_o$ candidate objects to obtain the feature representation of interacted object. 
Then we adopt a simple yet effective model $\mathcal{J}_O$, to integrate the re-weighted object information with the skeleton data for action classification.
We extend the node feature from only the coordinate information to the combined representation of human joint coordinate and its interacted object. From the skeleton flow to skeleton+object flow, it carries more semantic information and stronger expression ability.

\subsection{Loss Function}
\label{sec:loss}

In our framework, $SA$$\to$$OL$ and $SO$$\to$$AR$ are serialized  without sharing parameters across modules.
For human action recognition, we obtain two outputs from $\mathcal{F}_{S\to AR}$ and $\mathcal{F}_{SO\to AR}$ modules. We average the two outputs as our ultimate action classification results. Our loss function is the summation of the two action recognition losses $\mathcal{L}_{a1}$ and $\mathcal{L}_{a2}$ and the temporal consistency loss $\mathcal{L}_{con}$:
\begin{align}
    \mathcal{L} = \mathcal{L}_{a1} + \lambda_1\mathcal{L}_{a2} + \lambda_2\mathcal{L}_{con}
    \label{eq:loss_all}
\end{align}
$\lambda_1$ and $\lambda_2$ are scalars to weight the losses, $\mathcal{L}_{a1}$ and $\mathcal{L}_{a2}$ are both cross-entropy losses of action classification.

\section{Experiments}

\subsection{Datasets}


\noindent{\bf SYSU 3D Human-Object Interaction Dataset~\cite{sysu_dataset}.} This dataset contains 480 video clips of 12 activities performed by 40 actors. For each video clip, the RGB frames, depth sequences and skeleton data are captured by a Kinect camera. 
We adopt the same evaluation protocols as in~\cite{sysu_dataset}, \ieno, Cross Subject (CS) setting (half of the subjects for training and the rest for testing) and Same Subject (SS) setting (half of the samples of each subject for training and the rest for testing). For each setting, we report the mean accuracy results over 30 random splits as in~\cite{sysu_dataset}.

\noindent{\bf NTU60 RGB+D Dataset~\cite{ntudataset}.} NTU60 RGB+D is a large-scale and challenging human action dataset with 56,880 video clips together with the skeleton sequences of 60 actions performed by 40 subjects. The skeletons of each human is represented by 25 joints of 3D coordinates. In practice, we notice that NTU60 contains more than half non-interactive actions, such as ``clap'', ``salute'' and ``nod head/bow'', which is unfit for evaluating our model. Thus, we pick out 28 interactive actions to reorganize a new dataset with 26,368 video clips named \textbf{NTU60-HOI}. We have 17,789 video clips for training, 904 clips for valication, and 7,675 clips for testing.

We follow the previous experiment settings as~\cite{ntudataset,semantics2019} to evaluate our model. For the Cross Subject (CS) setting, half of the 40 subjects are selected for training. For the Cross View (CV) setting, the skeleton sequences captured by two cameras are used for training, and those captured by the third camera are used for testing. Following~\cite{ntudataset,semantics2019}, we randomly select 10\% for validation for both settings.

\noindent{\bf Northwestern-UCLA Dataset~\cite{ucla_dataset}.} It contains 1,494 video clips covering 10 actions performed by 10 actors. The RGB, depth and skeleton data are captured by three Kinect cameras simultaneously. There are three views for each action and each person has 20 3D joints.
We use the first two views for training and the third view for testing as in~\cite{ucla_dataset}.

\noindent{\bf UAV-Human Dataset~\cite{uav_human}.} UAV-Human is a newly proposed large-scale and comprehensive dataset with 67,428 multi-modal video sequences, 119 subjects and 22,476 frames for human behavior understanding. It contains diverse subjects, backgrounds, illuminations, weathers, occlusions, camera motions and UAV flying attitudes. For the task of skeleton-based action recognition, we follow the setting of Cross-Subject-V1 (CSV1) as in github repository~\cite{uav_github}.

\subsection{Implementation Details}
\noindent{\bf Data Processing.}
For a fair comparison, we adopt the same strategy to process the skeleton sequence as in~\cite{semantics2019}. For each video clip, the skeleton data is translated to be invariant to the initial positions based on the first frame. During the training, we down-sample the video sequence by equally segment the entire sequence into k clips and randomly select one frame in each clip to generate a new sequence, and k = 20, 20, 50, 50, 50 for NTU60 RGB+D, NTU60-HOI, SYSU-3D, NW-UCLA and UAV-Human datasets, respectively. During testing, similar to~\cite{baradel2018glimpse,semantics2019}, we randomly generate 5 down-sampled sequences as the input and take the mean score as our ultimate action classification results. During training, data augmentation is performed by randomly rotating the skeleton data by some degrees for robustness as same as in~\cite{semantics2019}. 

For the object detection part, we directly adopt the trained model from the github repository~\cite{ppdet2019} without fine-tuning.
We also take the correlation between the action types and object categories as a dataset-specific prior to filter out objects that can hardly be interacted with. Take the NTU60 RGB+D dataset as an example, for each action type, we pick out the interactable object classes from all the object classes based on our prior knowledge. Finally, we get an union of the selected object classes for all the action types of this dataset. Then the object classes outside of the union set are filtered/discarded.
We select $N_o=10$ objects with detection scores greater than 0.1 for all the three datasets.
For more than 10 objects, we select those with higher object detection scores; for less than 10 detected objects, we pad to 10 with the existing detected objects.
We also add a \textit{background} object category for those activities without interacted object.

\noindent{\bf Network Parameter Settings.}
For the $\mathcal{F}_{S\to AR}$ module, \ieno, the SGN model, we adopt the same network settings as in~\cite{semantics2019}. The number of neurons is set to 64 for $\Phi_1$ and $\Phi_2$ in Eq.~\ref{eq:sar1} with not shared parameters. And the dimension of the semantics of joint type $j_m$ is also 64. In Eq.~\ref{eq:sar2} of joint-level, $\theta_z(\cdot)$ and $\phi_z(\cdot)$ transformed the joint features to a 256-d embedding space. The dimensions of the three stacked GCN layers are 128, 256 and 256, respectively. In the frame-level module, the dimension of the first CNN layer is 256 with kernel size of 3, and the second CNN layer is 512-d with kernel size of 1. The number of neurons of the frame index semantics is 256.

For the $\mathcal{F}_{SA\to OL}$ module, $\Phi_j$ and $\Phi_a$ in Eq.~\ref{eq:saol1} have the dimension of 128.
For the $\mathcal{F}_{SO\to AR}$ module, the dimension of the interacted object and the skeleton data are both of dimension of 256. We choose $\lambda_1=2$ and $\lambda_2=1$ for the loss function since $\mathcal{F}_{SO\to AR}$ could be harder to learn than $\mathcal{F}_{S\to AR}$.

\noindent{\bf Training Procedure.}
Our implementations are based on the Pytorch framework with one P100 GPU. We apply the Adam optimizer with the initial learning rate of 0.001 for all the datasets. Similar to~\cite{semantics2019}, the learning rate decays by 1/10 at 60$^{th}$, 90$^{th}$ and 110$^{th}$ epoch, respectively. The training epochs are set to 120, 120, 100, 100, 120 and the batch sizes are set to 64, 64, 16, 16, 32 for NTU60 RGB+D, NTU60-HOI, SYSU-3D, NW-UCLA and UAV-Human, respectively. Following~\cite{semantics2019}, we utilize label smoothing with a factor of 0.1 and cross entropy loss for action classification to train the networks.

\begin{table}
\caption{The ablation study results for $SA\to OL$ and $SO\to AR$ modules on the SYSU-3D dataset in terms of accuracy (\%).}
\begin{center}
\begin{tabular}{cc|cc}
\hline
$SA\to OL$ & $SO\to AR$ & CS & SS\\
\hline\hline
$\times$ & $\times$  & 83.0 & 81.6\\
$\times$ & \checkmark & 86.5 & 84.4 \\
\checkmark & \checkmark & \textbf{88.5} & \textbf{87.5} \\
\hline
\end{tabular}
\end{center}
\label{tab:ablation}
\end{table}

\begin{table}
\caption{The ablation study results for the temporal consistency loss on the SYSU-3D dataset in terms of accuracy (\%).}
\begin{center}
\begin{tabular}{c|cc}
\hline
Consistency Loss & CS & SS\\
\hline\hline
$\times$ & 86.4 & 85.8 \\
\checkmark & \textbf{88.5} & \textbf{87.5} \\
\hline
\end{tabular}
\end{center}
\label{tab:ablation2}
\end{table}

\begin{table}[t]
\caption{The ablation study results for the parameters $\lambda_1$ and $\lambda_2$ on the SYSU-3D dataset in terms of accuracy (\%).}
\begin{center}
\begin{tabular}{cc|c}
\hline
$\lambda_1$ & $\lambda_2$ & Accuracy\\
\hline\hline
1 & 1 & 88.33\\
2 & 1 & \textbf{88.54}\\
3 & 1 & 88.49\\
2 & 2 & 88.07\\
\hline
\end{tabular}
\end{center}
\label{tab:lambda}
\end{table}

\subsection{Ablation Study}
We evaluate the effectiveness of each component of our framework extensively on SYSU-3D dataset. The mean action classification accuracy results over 30 random splits on the Cross-Subject (CS) and Same-Subject (SS) setting of SYSU-3D dataset are reported. We choose SYSU-3D because this dataset is designed for studying human object interactions in videos and it is challenging with numerous background objects for interacted object localization. Note that, we only use joint data for training without any pretraining on other datasets.



\subsubsection{Effectiveness of joint learning}
To demonstrate the effectiveness of our joint task learning framework, we perform an ablation experiment on two main modules: $SA\to OL$ and $SO\to AR$ (in Tab.~\ref{tab:ablation}). We have two main observations:
\begin{itemize}[leftmargin=*,noitemsep,nolistsep]
    \item[1)] By integrating object information with skeleton data, \ieno, $SO\to AR$, the performance improves by 3.5\% and 2.8\% for the CS and SS settings, respectively.
    \item[2)] By incorporating action type to assist interacted object localization, \ieno, $SA\to OL$, the performance improves by 2.0\% and 3.1\% for the CS and SS settings, respectively.
\end{itemize}
In summary, we can derive that interacted object localization and human action recognition can mutually promote each other in the joint learning framework.

\subsubsection{Effectiveness of temporal consistency loss}
We design two experiments (in Tab.~\ref{tab:ablation2}) to validate the effectiveness of temporal consistency loss introduced in Sect.~\ref{sec:sa_ol}.
We have the observation that with consistency loss, the accuracy improves by 2.1\% and 1.7\% for the CS and SS settings, respectively. This verifies that employing the temporal consistency of interacted object categories is beneficial for localizing more accurate objects and in turn helps human action recognition.

\subsubsection{Parameters of $\lambda_1$ and $\lambda_2$}
\label{append:lambda} We provide the performances on the CS setting of SYSU-3D dataset for different settings of $\lambda_1$ and $\lambda_2$ in Tab.~\ref{tab:lambda}. We can observe that $\lambda_1$=2, $\lambda_2$=1 yields the best performance.

\begin{table}
\caption{Comparisons of the accuracy (\%) on the SYSU-3D dataset with state-of-the-art methods.}
\begin{center}
\begin{tabular}{l|ccc}
\hline
Methods & Year & CS & SS \\
\hline\hline
VA-LSTM~\cite{viewadaptive} & 2017 & 77.5 & 76.9\\
ST-LSTM~\cite{liu2017skeleton} & 2018 & 76.5 & -\\
GR-GCN~\cite{gao2019optimized} & 2019 & 77.9 & -\\
Two stream GCA-LSTM~\cite{liu2017skeletonbase} & 2017 & 77.9 & - \\
SR-TSL~\cite{si2018skeleton} & 2018 & 81.9 & 80.7 \\
\hline\hline
Baseline(SGN) & 2020 & 83.0 & 81.6 \\
Ours & - & \textbf{88.5} & \textbf{87.5} \\
\hline
\end{tabular}
\end{center}
\label{tab:sysu}
\end{table}

\begin{table}
\caption{Comparisons of the accuracy (\%) on the NTU60-HOI dataset with state-of-the-art methods. $^\dagger$ denotes the result is reproduced by us.}
\begin{center}
\begin{tabular}{l|ccc}
\hline
Methods & Year & CS & CV \\
\hline\hline
2s-AGCN$^\dagger$~\cite{two-stream} & 2019 & 84.1 & 92.4 \\
1s Shift-GCN$^\dagger$~\cite{cheng2020skeleton} & 2020 & 84.5 & 92.8 \\
MS-G3D Net$^\dagger$~\cite{liu2020disentangling} & 2020 & 85.7 & 92.9 \\
\hline\hline
Baseline(SGN)$^\dagger$ & 2020 & 85.3 & 92.4 \\
Ours & - & \textbf{87.7} & \textbf{94.8} \\
\hline
\end{tabular}
\end{center}
\label{tab:ntu}
\end{table}

\begin{table}[ht]
\caption{Comparisons of the accuracy (\%) on the full NTU60 RGB+D dataset with state-of-the-art methods.}
\begin{center}
\begin{tabular}{l|ccc}
\hline
Methods & Year & CS & CV \\
\hline\hline
Lie Group~\cite{veeriah2015differential} & 2015 & 50.1 & 52.8 \\
HBRNN-L~\cite{du2015hierarchical} & 2015 & 59.1 & 64.0 \\
Part-Aware LSTM~\cite{ntudataset} & 2016 & 62.9 & 70.3 \\
ST-LSTM + Trust Gate~\cite{liu2016spatio} & 2016 & 69.2 & 77.7\\
STA-LSTM~\cite{song2016end} & 2017 & 73.4 & 81.2\\ 
GCA-LSTM~\cite{liu2017global} & 2017 & 74.4 & 82.8\\
Clips+CNN+MTLN~\cite{ke2017new} & 2017 & 79.6 & 84.8 \\
VA-LSTM~\cite{viewadaptive} & 2017 & 79.4 & 87.6\\
ElAtt-GRU~\cite{zhang2018adding} & 2018 &80.7& 88.4 \\
ST-GCN~\cite{st-gcn} & 2018 & 81.5 & 88.3\\
DPRL+GCNN~\cite{tang2018deep} & 2018 & 83.5 & 89.8\\
SR-TSL~\cite{si2018skeleton} & 2018 & 84.8 & 92.4\\
HCN~\cite{li2018co} & 2018 & 86.5 & 91.1 \\
AGC-LSTM (joint)~\cite{si2019attention} & 2019 & 87.5 & 93.5\\
AS-GCN~\cite{li2019actional} & 2019 & 86.8 & 94.2 \\
GR-GCN~\cite{gao2019optimized} & 2019 & 87.5 &  94.3 \\
2s-AGCN~\cite{two-stream} & 2019 & 88.5 & 95.1 \\
VA-CNN~\cite{zhang2019view} & 2019 & 88.7 & 94.3 \\
SGN~\cite{semantics2019} & 2020 & 89.0 & 94.5 \\
Advanced CA-GCN~\cite{zhang2020context} & 2020 & 83.5 & 91.4 \\
1s Shift-GCN~\cite{cheng2020skeleton} & 2020 & 87.8 & 95.1 \\
2s Shift-GCN~\cite{cheng2020skeleton} & 2020 & 89.7 & 96.0 \\
4s Shift-GCN~\cite{cheng2020skeleton} & 2020 & 90.7 & 96.5 \\
MS-G3D Net~\cite{liu2020disentangling} & 2020 & 91.5 & 96.2 \\
DC-GCN+ADG~\cite{chengdecoupling} & 2020 & 90.8 & 96.6 \\
CTR-GCN~\cite{chen2021channel} & 2021 & \textbf{92.4} & \textbf{96.8} \\
\hline\hline
STA-Hands~\cite{baradel2017human} & 2017 & 82.5 & 88.6\\
altered STA-Hands~\cite{baradel2018human1} & 2018 & 84.8 & 90.6 \\
Glimpse Cloud~\cite{baradel2018glimpse} & 2018 & 86.6 & 93.2 \\
PEM~\cite{liu2018recognizing} & 2018 & 91.7 & 95.2 \\
SI-MM (RGB+Ske)~\cite{song2018skeleton} & 2018 & 85.1 & 92.8 \\
SI-MM (RGB+Ske+Flow)~\cite{song2018skeleton} & 2018 & 92.6 & 97.9 \\
Separable STA~\cite{das2019toyota} & 2019 & 92.2 & 94.6 \\
P-I3D~\cite{wheretofocus} & 2020 & 93.0 & 95.4 \\
Cross-Attention~\cite{fan2020context} & 2020 & 84.2 & 89.3 \\
VPN~\cite{das2020vpn} & 2020 & \textbf{95.5} & \textbf{98.0} \\
\hline\hline
Baseline (SGN~\cite{semantics2019}) & 2020 & 89.0 & 94.5 \\
Ours & - & 90.0 & 95.7\\
\hline
\end{tabular}
\end{center}
\label{tab:ntu_full}
\end{table}

\subsection{Visualization of Interacted Object Localization}
\begin{figure*}
\begin{center}
  \includegraphics[width=1.0\linewidth]{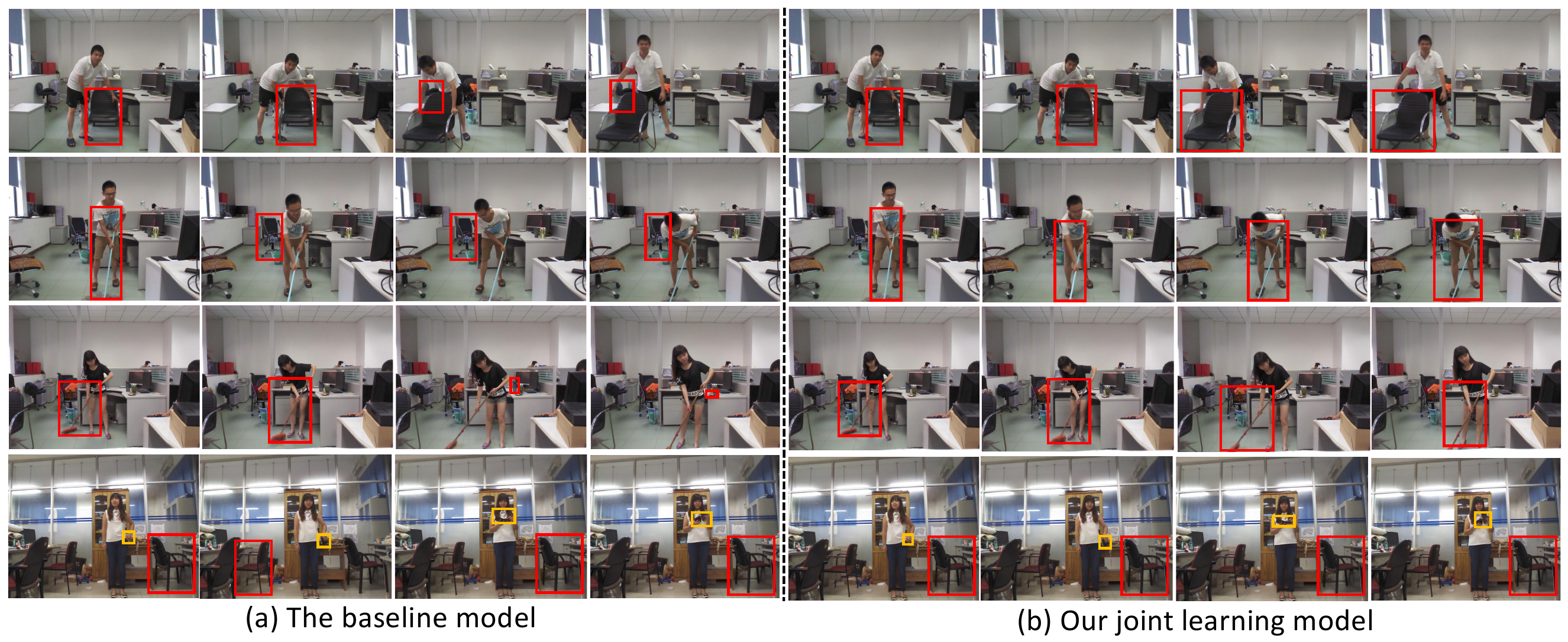}
\end{center}
  \caption{Visualization results of the interacted object localization. The left rows are from the baseline model and the right rows are from ours. The \tcr{red} boxes indicate the estimated interacted object, the \textcolor{yellow}{yellow} boxes on the last row indicate the undetected interacted object. The four rows are selected from Jianshen/video8, luxi/video11, Yumeng/video12 and liwuyang/video10 of the SYSU-3D dataset, respectively.}
\label{fig:visualize}
\end{figure*}

Due to the absence of ground-truth interacted human-object bounding boxes, we only provide the qualitative visualization results of the baseline model and our joint learning model in Fig.~\ref{fig:visualize}. The differences between these two models lie in the $SA\to OL$ module and the temporal consistency loss. We observe that the joint learning framework with temporal consistency loss provides more reasonable and robust interacted object localization results over the baseline method.

\begin{table}
\caption{Comparisons of the accuracy (\%) on the NW-UCLA dataset with state-of-the-art methods.}
\begin{center}
\begin{tabular}{l|ccc}
\hline
Methods & Year & Accuracy \\
\hline\hline
Lie Group~\cite{veeriah2015differential} & 2014 & 74.2 \\
Actionlet ensemble~\cite{wang2013learning} & 2014 & 76.0 \\
Visualization CNN~\cite{liu2017enhanced} & 2017 & 86.1 \\
Ensemble TS-LSTM~\cite{lee2017ensemble} & 2017 & 89.2 \\
2s AGC-LSTM~\cite{si2019attention} & 2019 & 93.3 \\
4s Shift-GCN~\cite{cheng2020skeleton} & 2020 & 94.6 \\
DC-GCN+ADG~\cite{chengdecoupling} & 2020 & 95.3 \\
CTR-GCN~\cite{chen2021channel} & 2021 & 96.5 \\
\hline\hline
NKTM~\cite{rahmani2015learning} & 2015 & 85.6 \\
HPM+TM~\cite{rahmani20163d} & 2016 & 91.0 \\
Glimpse Cloud~\cite{baradel2018glimpse} & 2018 & 90.1 \\
Separable STA~\cite{das2019toyota} & 2019 & 92.4 \\
P-I3D~\cite{wheretofocus} & 2019 & 93.1 \\
VPN~\cite{das2020vpn} & 2020 & 93.5 \\
\hline\hline
Baseline(SGN)-joint & 2020 & 92.5 \\
Ours-joint & - & 95.7 \\
Ours-bone & - & 94.6 \\
Ours-2stream & - & \textbf{97.2} \\
\hline
\end{tabular}
\end{center}
\label{tab:ucla}
\end{table}

\begin{table}
\caption{The ablation study results for the temporal consistency loss on the full UAV-Human dataset in terms of accuracy (\%).}
\begin{center}
\begin{tabular}{c|cc}
\hline
Methods & Year & Accuracy\\
\hline\hline
DGNN~\cite{shi2019skeleton} & 2019 & 29.90 \\
ST-GCN~\cite{st-gcn} & 2018 & 30.25\\
2s-AGCN~\cite{two-stream} & 2019 & 34.84\\
HARD-Net~\cite{li2020hard} & 2020 & 36.97\\
Shift-GCN~\cite{cheng2020skeleton} & 2020 & 37.98\\
\hline\hline
Baseline(SGN) & 2020 & 36.37\\
Ours & - & \textbf{39.99} \\
\hline
\end{tabular}
\end{center}
\label{tab:uav}
\end{table}

The action type of the four rows in Fig.~\ref{fig:visualize} are ``moving chair'', ``mopping'', ``sweeping'' and ``taking from wallet''. The upper three rows show that our joint learning framework obtains more consistent results, while the baseline model may change the interacted object along with the person movement. We also visualize the cases in the last row that the wallet is undetected, both the baseline method and ours fail. It inspires us to exploit a more general and robust interacted object detector in videos in future works.

\subsection{Comparisons with the State-of-the-arts}
We compare our joint learning framework with other state-of-the-art models on SYSU-3D, NTU60-HOI, full NTU60 RGB+D, NW-UCLA and UAV-Human datasets in Tab.~\ref{tab:sysu}, Tab.~\ref{tab:ntu}, Tab.~\ref{tab:ntu_full} and Tab.~\ref{tab:ucla}, respectively. 

\subsubsection{SYSU-3D}
SYSU-3D is focusing on human-object interactions in videos. As shown in Tab.~\ref{tab:sysu}, we only utilize the joint feature without any pretraining on NTU-60 dataset for a fair comparison with the baseline model SGN~\cite{semantics2019}. Our proposed joint learning framework outperforms the baseline method by 5.5\% and 5.9\% for the CS and SS settings and sets up a new state-of-the-art.

\subsubsection{NTU60-HOI}
NTU60 RGB+D is a large scale human activity dataset and we construct a subset dataset NTU60-HOI with only the interactive actions. We treat SGN~\cite{semantics2019} as a strong baseline and show the comparison in Tab.~\ref{tab:ntu}. Our framework outperforms the baseline by 2.4\% and 2.4\% for the CS and CV settings, respectively. We also re-produce the results of some other latest approaches~\cite{two-stream,cheng2020skeleton,liu2020disentangling} using their provided source code. We can see that our method achieves the state-of-the-art performance. It confirms the effectiveness of our method on large-scale video action benchmarks.

\subsubsection{NTU60 RGB+D}
For the full NTU60 RGB+D dataset, we present the comparisons in Tab.~\ref{tab:ntu_full}.
The methods in the \textbf{upper rows} belong to the skeleton-only based methods. Particularly, the 4s Shift-GCN~\cite{cheng2020skeleton}, DC-GCN+ADG~\cite{chengdecoupling} and CTR-GCN~\cite{chen2021channel} models ensemble the results of 4 streams: joint, bone, joint motion and bone motion while we only use the joint stream as in~\cite{semantics2019}. MS-G3D Net~\cite{liu2020disentangling} leverages dense cross space-time information propagation with high computational complexity.
In the \textbf{middle rows}, we present those methods using RGB information or multi-modality features. Particularly, PEM~\cite{liu2018recognizing} takes the estimated pose heatmap from RGB frames as input for action prediction. SI-MM (RGB+Ske)~\cite{song2018skeleton} fuses the RGB information and the skeleton in multiple streams and SI-MM (RGB+Ske+Flow) uses extra optical flow data. Separable STA~\cite{das2019toyota}, P-I3D~\cite{wheretofocus}, Cross-Attention~\cite{fan2020context}, and VPN~\cite{das2020vpn} all design sophisticated visual backbone networks (\ieno, I3D~\cite{carreira2017quo}) pre-trained on the very large-scale dataset Kinetics-400~\cite{kay2017kinetics}.
The \textbf{bottom two rows} show the performance of the baseline and our proposed method. Our method outperforms the baseline method by 1.0\% and 1.2\% for the CS and CV settings, respectively.

\subsubsection{Northwestern-UCLA}
For the NW-UCLA dataset, we choose the SGN~\cite{semantics2019} as our baseline and implement our joint learning framework based on it. 
In Tab.~\ref{tab:ucla}, the methods in the \textbf{upper rows} only use skeleton data. 
The methods in the \textbf{middle rows} utilize RGB information or multi-modality information. Particularly, separable STA~\cite{das2019toyota}, P-I3D~\cite{wheretofocus}, and VPN~\cite{das2020vpn} all design sophisticated visual backbone networks pre-trained on the very large-scale dataset Kinetics-400~\cite{kay2017kinetics}. 
In the \textbf{bottom rows}, ``joint'' means only utilizing skeleton joints data, ``bone'' means only utilizing the bones (physical connections between joints) feature as introduced in~\cite{two-stream} and ``2stream'' means the fusion of joint-stream and bone-stream as in~\cite{two-stream}. Note that, DC-GCN+ADG~\cite{chengdecoupling} ensembles 4 streams of joint, bone, joint motion and bone motion as in~\cite{shi2019skeleton}. Since~\cite{semantics2019} implicitly calculates the velocity of joints similar to the joint/bone motion design, we only report the fusion result of joint-stream and bone-stream for a fair comparison. Our framework brings gains of 3.2\% for the joint stream, 4.7\% for the fusion stream over the baseline model and sets up a new state-of-the-art.

\subsubsection{UAV-Human}
For the full UAV-Human dataset, the comparisons of skeleton-based action recognition results are presented in Tab.~\ref{tab:uav}. The UAV-Human dataset is challenging with variations of backgrounds, occlusions, illuminations, camera motions, \etcno. The \textbf{bottom two rows} show that our method outperforms the baseline method by 3.62\% and sets up a new state-of-the-art.

\section{Limitations and Future Works}
In this work, we verify the effectiveness of the joint learning framework for interacted object localization and human action recognition using skeleton data. We discuss some limitations and future works here.

\begin{itemize}[leftmargin=*,noitemsep,nolistsep]
    \item Our implementation for interacted object localization is based on the preliminary object detection results. The failure of object detection will hurt the subsequent interacted object localization. Thus a more general and robust interacted object detector in videos is desired.
    \item Our framework can be extended to other modalities, such as RGB/RGB+D video sequences. We can design visual backbone networks like I3D~\cite{carreira2017quo} to directly leverage the RGB/RGB+D data. The motion information between frames can be better exploited for localizing interacted object.
    \item By integrating interacted object with the skeleton data, we extend the pose flow to the pose+object flow. The \textit{semantics} of human joints interacting with the objects could be better explored.
\end{itemize}

\section{Conclusion}

In this work, a joint task learning framework for skeleton-based ``interacted object localization'' and ``human action recognition'' without ground-truth of interacted object is proposed.
The two tasks are serialized together and mutually assist each other in a collaborative manner.
To make up for the absence of labels of interacted human-object boxes, we introduce the temporal consistency loss to strengthen the interacted object localization. Quantitative and qualitative results show that our joint learning framework outperforms the baseline models for action recognition by a large margin and also achieves reasonable interacted object localization results. We hope that our work will inspire more future works on interacted object localization and human action recognition.

\ifCLASSOPTIONcaptionsoff
  \newpage
\fi

\bibliographystyle{IEEEtran}
\bibliography{actionso_arxiv}

\begin{thebibliography}{100}
\providecommand{\url}[1]{#1}
\csname url@samestyle\endcsname
\providecommand{\newblock}{\relax}
\providecommand{\bibinfo}[2]{#2}
\providecommand{\BIBentrySTDinterwordspacing}{\spaceskip=0pt\relax}
\providecommand{\BIBentryALTinterwordstretchfactor}{4}
\providecommand{\BIBentryALTinterwordspacing}{\spaceskip=\fontdimen2\font plus
\BIBentryALTinterwordstretchfactor\fontdimen3\font minus
  \fontdimen4\font\relax}
\providecommand{\BIBforeignlanguage}[2]{{%
\expandafter\ifx\csname l@#1\endcsname\relax
\typeout{** WARNING: IEEEtran.bst: No hyphenation pattern has been}%
\typeout{** loaded for the language `#1'. Using the pattern for}%
\typeout{** the default language instead.}%
\else
\language=\csname l@#1\endcsname
\fi
#2}}
\providecommand{\BIBdecl}{\relax}
\BIBdecl

\bibitem{poppe2010survey}
R.~Poppe, ``A survey on vision-based human action recognition,'' \emph{Image
  and vision computing}, vol.~28, no.~6, pp. 976--990, 2010.

\bibitem{weinland2011survey}
D.~Weinland, R.~Ronfard, and E.~Boyer, ``A survey of vision-based methods for
  action representation, segmentation and recognition,'' \emph{Computer vision
  and image understanding}, vol. 115, no.~2, pp. 224--241, 2011.

\bibitem{aggarwalsurvey}
J.~K. Aggarwal and M.~S. Ryoo, ``Human activity analysis: A review,'' \emph{ACM
  Computing Surveys (CSUR)}, vol.~43, no.~3, pp. 1--43, 2011.

\bibitem{st-gcn}
S.~Yan, Y.~Xiong, and D.~Lin, ``Spatial temporal graph convolutional networks
  for skeleton-based action recognition,'' \emph{arXiv preprint
  arXiv:1801.07455}, 2018.

\bibitem{two-stream}
L.~Shi, Y.~Zhang, J.~Cheng, and H.~Lu, ``Two-stream adaptive graph
  convolutional networks for skeleton-based action recognition,'' in
  \emph{CVPR}, 2019, pp. 12\,026--12\,035.

\bibitem{li2019actional}
M.~Li, S.~Chen, X.~Chen, Y.~Zhang, Y.~Wang, and Q.~Tian, ``Actional-structural
  graph convolutional networks for skeleton-based action recognition,'' in
  \emph{CVPR}, 2019, pp. 3595--3603.

\bibitem{semantics2019}
P.~Zhang, C.~Lan, W.~Zeng, J.~Xing, J.~Xue, and N.~Zheng, ``Semantics-guided
  neural networks for efficient skeleton-based human action recognition,'' in
  \emph{CVPR}, 2020, pp. 1112--1121.

\bibitem{zhang2020context}
X.~Zhang, C.~Xu, and D.~Tao, ``Context aware graph convolution for
  skeleton-based action recognition,'' in \emph{CVPR}, 2020, pp.
  14\,333--14\,342.

\bibitem{cheng2020skeleton}
K.~Cheng, Y.~Zhang, X.~He, W.~Chen, J.~Cheng, and H.~Lu, ``Skeleton-based
  action recognition with shift graph convolutional network,'' in \emph{CVPR},
  2020, pp. 183--192.

\bibitem{liu2020disentangling}
Z.~Liu, H.~Zhang, Z.~Chen, Z.~Wang, and W.~Ouyang, ``Disentangling and unifying
  graph convolutions for skeleton-based action recognition,'' in \emph{CVPR},
  2020, pp. 143--152.

\bibitem{su2020predict}
K.~Su, X.~Liu, and E.~Shlizerman, ``Predict \& cluster: Unsupervised skeleton
  based action recognition,'' in \emph{CVPR}, 2020, pp. 9631--9640.

\bibitem{chengdecoupling}
K.~Cheng, Y.~Zhang, C.~Cao, L.~Shi, J.~Cheng, and H.~Lu, ``Decoupling gcn with
  dropgraph module for skeleton-based action recognition,'' in \emph{ECCV},
  2020.

\bibitem{zhang2020deep}
T.~Zhang, W.~Zheng, Z.~Cui, Y.~Zong, C.~Li, X.~Zhou, and J.~Yang, ``Deep
  manifold-to-manifold transforming network for skeleton-based action
  recognition,'' \emph{IEEE Transactions on Multimedia}, vol.~22, no.~11, pp.
  2926--2937, 2020.

\bibitem{zhu2019cuboid}
K.~Zhu, R.~Wang, Q.~Zhao, J.~Cheng, and D.~Tao, ``A cuboid cnn model with an
  attention mechanism for skeleton-based action recognition,'' \emph{IEEE
  Transactions on Multimedia}, vol.~22, no.~11, pp. 2977--2989, 2019.

\bibitem{hu2019joint}
G.~Hu, B.~Cui, and S.~Yu, ``Joint learning in the spatio-temporal and frequency
  domains for skeleton-based action recognition,'' \emph{IEEE Transactions on
  Multimedia}, vol.~22, no.~9, pp. 2207--2220, 2019.

\bibitem{avola20192}
D.~Avola, M.~Cascio, L.~Cinque, G.~L. Foresti, C.~Massaroni, and E.~Rodol{\`a},
  ``2-d skeleton-based action recognition via two-branch stacked lstm-rnns,''
  \emph{IEEE Transactions on Multimedia}, vol.~22, no.~10, pp. 2481--2496,
  2019.

\bibitem{liu2020multi}
K.~Liu, L.~Gao, N.~M. Khan, L.~Qi, and L.~Guan, ``A multi-stream graph
  convolutional networks-hidden conditional random field model for
  skeleton-based action recognition,'' \emph{IEEE Transactions on Multimedia},
  vol.~23, pp. 64--76, 2020.

\bibitem{yang2020hierarchical}
J.~Yang, W.~Liu, J.~Yuan, and T.~Mei, ``Hierarchical soft quantization for
  skeleton-based human action recognition,'' \emph{IEEE Transactions on
  Multimedia}, vol.~23, pp. 883--898, 2020.

\bibitem{something-else}
J.~Materzynska, T.~Xiao, R.~Herzig, H.~Xu, X.~Wang, and T.~Darrell,
  ``Something-else: Compositional action recognition with spatial-temporal
  interaction networks,'' in \emph{CVPR}, 2020, pp. 1049--1059.

\bibitem{tang2020asynchronous}
J.~Tang, J.~Xia, X.~Mu, B.~Pang, and C.~Lu, ``Asynchronous interaction
  aggregation for action detection,'' \emph{arXiv preprint arXiv:2004.07485},
  2020.

\bibitem{pastanet}
Y.-L. Li, L.~Xu, X.~Liu, X.~Huang, Y.~Xu, S.~Wang, H.-S. Fang, Z.~Ma, M.~Chen,
  and C.~Lu, ``Pastanet: Toward human activity knowledge engine,'' in
  \emph{CVPR}, 2020, pp. 382--391.

\bibitem{gkioxari2018detecting}
G.~Gkioxari, R.~Girshick, P.~Doll{\'a}r, and K.~He, ``Detecting and recognizing
  human-object interactions,'' in \emph{CVPR}, 2018, pp. 8359--8367.

\bibitem{sysu_dataset}
J.-F. Hu, W.-S. Zheng, J.~Lai, and J.~Zhang, ``Jointly learning heterogeneous
  features for rgb-d activity recognition,'' in \emph{CVPR}, 2015, pp.
  5344--5352.

\bibitem{ntudataset}
A.~Shahroudy, J.~Liu, T.-T. Ng, and G.~Wang, ``Ntu rgb+ d: A large scale
  dataset for 3d human activity analysis,'' in \emph{CVPR}, 2016, pp.
  1010--1019.

\bibitem{ucla_dataset}
J.~Wang, X.~Nie, Y.~Xia, Y.~Wu, and S.-C. Zhu, ``Cross-view action modeling,
  learning and recognition,'' in \emph{CVPR}, 2014, pp. 2649--2656.

\bibitem{uav_human}
T.~Li, J.~Liu, W.~Zhang, Y.~Ni, W.~Wang, and Z.~Li, ``Uav-human: A large
  benchmark for human behavior understanding with unmanned aerial vehicles,''
  in \emph{CVPR}, 2021, pp. 16\,266--16\,275.

\bibitem{ji20123d}
S.~Ji, W.~Xu, M.~Yang, and K.~Yu, ``3d convolutional neural networks for human
  action recognition,'' \emph{TPAMI}, vol.~35, no.~1, pp. 221--231, 2012.

\bibitem{tran2014c3d}
D.~Tran, L.~D. Bourdev, R.~Fergus, L.~Torresani, and M.~Paluri, ``C3d: generic
  features for video analysis,'' \emph{CoRR, abs/1412.0767}, vol.~2, no.~7,
  p.~8, 2014.

\bibitem{taylor2010convolutional}
G.~W. Taylor, R.~Fergus, Y.~LeCun, and C.~Bregler, ``Convolutional learning of
  spatio-temporal features,'' in \emph{ECCV}.\hskip 1em plus 0.5em minus
  0.4em\relax Springer, 2010, pp. 140--153.

\bibitem{varol2017long}
G.~Varol, I.~Laptev, and C.~Schmid, ``Long-term temporal convolutions for
  action recognition,'' \emph{TPAMI}, vol.~40, no.~6, pp. 1510--1517, 2017.

\bibitem{carreira2017quo}
J.~Carreira and A.~Zisserman, ``Quo vadis, action recognition? a new model and
  the kinetics dataset,'' in \emph{CVPR}, 2017, pp. 6299--6308.

\bibitem{qiu2017learning}
Z.~Qiu, T.~Yao, and T.~Mei, ``Learning spatio-temporal representation with
  pseudo-3d residual networks,'' in \emph{ICCV}, 2017, pp. 5533--5541.

\bibitem{diba2018spatio}
A.~Diba, M.~Fayyaz, V.~Sharma, M.~Mahdi~Arzani, R.~Yousefzadeh, J.~Gall, and
  L.~Van~Gool, ``Spatio-temporal channel correlation networks for action
  classification,'' in \emph{ECCV}, 2018, pp. 284--299.

\bibitem{tran2018closer}
D.~Tran, H.~Wang, L.~Torresani, J.~Ray, Y.~LeCun, and M.~Paluri, ``A closer
  look at spatiotemporal convolutions for action recognition,'' in \emph{CVPR},
  2018, pp. 6450--6459.

\bibitem{xie2017rethinking}
S.~Xie, C.~Sun, J.~Huang, Z.~Tu, and K.~Murphy, ``Rethinking spatiotemporal
  feature learning for video understanding,'' \emph{arXiv preprint
  arXiv:1712.04851}, vol.~1, no.~2, p.~5, 2017.

\bibitem{christoph2016spatiotemporal}
R.~Christoph and F.~A. Pinz, ``Spatiotemporal residual networks for video
  action recognition,'' \emph{Neurips}, pp. 3468--3476, 2016.

\bibitem{soomro2012ucf101}
K.~Soomro, A.~R. Zamir, and M.~Shah, ``Ucf101: A dataset of 101 human actions
  classes from videos in the wild,'' \emph{arXiv preprint arXiv:1212.0402},
  2012.

\bibitem{kay2017kinetics}
W.~Kay, J.~Carreira, K.~Simonyan, B.~Zhang, C.~Hillier, S.~Vijayanarasimhan,
  F.~Viola, T.~Green, T.~Back, P.~Natsev \emph{et~al.}, ``The kinetics human
  action video dataset,'' \emph{arXiv preprint arXiv:1705.06950}, 2017.

\bibitem{gu2018ava}
C.~Gu, C.~Sun, D.~A. Ross, C.~Vondrick, C.~Pantofaru, Y.~Li,
  S.~Vijayanarasimhan, G.~Toderici, S.~Ricco, R.~Sukthankar \emph{et~al.},
  ``Ava: A video dataset of spatio-temporally localized atomic visual
  actions,'' in \emph{CVPR}, 2018, pp. 6047--6056.

\bibitem{caba2015activitynet}
B.~G. Fabian Caba~Heilbron, Victor~Escorcia and J.~C. Niebles, ``Activitynet: A
  large-scale video benchmark for human activity understanding,'' in
  \emph{CVPR}, 2015, pp. 961--970.

\bibitem{gupta2020quo}
P.~Gupta, A.~Thatipelli, A.~Aggarwal, S.~Maheshwari, N.~Trivedi, S.~Das, and
  R.~K. Sarvadevabhatla, ``Quo vadis, skeleton action recognition ?'' 2020.

\bibitem{feichtenhofer2019slowfast}
C.~Feichtenhofer, H.~Fan, J.~Malik, and K.~He, ``Slowfast networks for video
  recognition,'' in \emph{CVPR}, 2019, pp. 6202--6211.

\bibitem{hou2017tube}
R.~Hou, C.~Chen, and M.~Shah, ``Tube convolutional neural network (t-cnn) for
  action detection in videos,'' in \emph{ICCV}, 2017, pp. 5822--5831.

\bibitem{hussein2013human}
M.~E. Hussein, M.~Torki, M.~A. Gowayyed, and M.~El-Saban, ``Human action
  recognition using a temporal hierarchy of covariance descriptors on 3d joint
  locations,'' in \emph{IJCAI}, 2013.

\bibitem{wang2012mining}
J.~Wang, Z.~Liu, Y.~Wu, and J.~Yuan, ``Mining actionlet ensemble for action
  recognition with depth cameras,'' in \emph{CVPR}.\hskip 1em plus 0.5em minus
  0.4em\relax IEEE, 2012, pp. 1290--1297.

\bibitem{vemulapalli2014human}
R.~Vemulapalli, F.~Arrate, and R.~Chellappa, ``Human action recognition by
  representing 3d skeletons as points in a lie group,'' in \emph{CVPR}, 2014,
  pp. 588--595.

\bibitem{du2015hierarchical}
Y.~Du, W.~Wang, and L.~Wang, ``Hierarchical recurrent neural network for
  skeleton based action recognition,'' in \emph{CVPR}, 2015, pp. 1110--1118.

\bibitem{liu2016spatio}
J.~Liu, A.~Shahroudy, D.~Xu, and G.~Wang, ``Spatio-temporal lstm with trust
  gates for 3d human action recognition,'' in \emph{ECCV}.\hskip 1em plus 0.5em
  minus 0.4em\relax Springer, 2016, pp. 816--833.

\bibitem{song2016end}
S.~Song, C.~Lan, J.~Xing, W.~Zeng, and J.~Liu, ``An end-to-end spatio-temporal
  attention model for human action recognition from skeleton data,''
  \emph{arXiv preprint arXiv:1611.06067}, 2016.

\bibitem{viewadaptive}
P.~Zhang, C.~Lan, J.~Xing, W.~Zeng, J.~Xue, and N.~Zheng, ``View adaptive
  recurrent neural networks for high performance human action recognition from
  skeleton data,'' in \emph{ICCV}, 2017, pp. 2117--2126.

\bibitem{li2018independently}
S.~Li, W.~Li, C.~Cook, C.~Zhu, and Y.~Gao, ``Independently recurrent neural
  network (indrnn): Building a longer and deeper rnn,'' in \emph{CVPR}, 2018,
  pp. 5457--5466.

\bibitem{cao2018skeleton}
C.~Cao, C.~Lan, Y.~Zhang, W.~Zeng, H.~Lu, and Y.~Zhang, ``Skeleton-based action
  recognition with gated convolutional neural networks,'' \emph{TCSVT},
  vol.~29, no.~11, pp. 3247--3257, 2018.

\bibitem{liu2017two}
H.~Liu, J.~Tu, and M.~Liu, ``Two-stream 3d convolutional neural network for
  skeleton-based action recognition,'' \emph{arXiv preprint arXiv:1705.08106},
  2017.

\bibitem{kim2017interpretable}
T.~S. Kim and A.~Reiter, ``Interpretable 3d human action analysis with temporal
  convolutional networks,'' in \emph{CVPRW}.\hskip 1em plus 0.5em minus
  0.4em\relax IEEE, 2017, pp. 1623--1631.

\bibitem{ke2017new}
Q.~Ke, M.~Bennamoun, S.~An, F.~Sohel, and F.~Boussaid, ``A new representation
  of skeleton sequences for 3d action recognition,'' in \emph{CVPR}, 2017, pp.
  3288--3297.

\bibitem{liu2017enhanced}
M.~Liu, H.~Liu, and C.~Chen, ``Enhanced skeleton visualization for view
  invariant human action recognition,'' \emph{Pattern Recognition}, vol.~68,
  pp. 346--362, 2017.

\bibitem{li2017skeleton}
C.~Li, Q.~Zhong, D.~Xie, and S.~Pu, ``Skeleton-based action recognition with
  convolutional neural networks,'' in \emph{ICMEW}.\hskip 1em plus 0.5em minus
  0.4em\relax IEEE, 2017, pp. 597--600.

\bibitem{liu2018recognizing}
M.~Liu and J.~Yuan, ``Recognizing human actions as the evolution of pose
  estimation maps,'' in \emph{CVPR}, 2018, pp. 1159--1168.

\bibitem{shi2019skeleton}
L.~Shi, Y.~Zhang, J.~Cheng, and H.~Lu, ``Skeleton-based action recognition with
  directed graph neural networks,'' in \emph{CVPR}, 2019, pp. 7912--7921.

\bibitem{chen2021channel}
Y.~Chen, Z.~Zhang, C.~Yuan, B.~Li, Y.~Deng, and W.~Hu, ``Channel-wise topology
  refinement graph convolution for skeleton-based action recognition,'' in
  \emph{Proceedings of the IEEE/CVF International Conference on Computer
  Vision}, 2021, pp. 13\,359--13\,368.

\bibitem{Lin_2020}
\BIBentryALTinterwordspacing
L.~Lin, S.~Song, W.~Yang, and J.~Liu, ``Ms2l,'' \emph{ACM Multimedia}, Oct
  2020. [Online]. Available: \url{http://dx.doi.org/10.1145/3394171.3413548}
\BIBentrySTDinterwordspacing

\bibitem{li2019transferable}
Y.-L. Li, S.~Zhou, X.~Huang, L.~Xu, Z.~Ma, H.-S. Fang, Y.~Wang, and C.~Lu,
  ``Transferable interactiveness knowledge for human-object interaction
  detection,'' in \emph{CVPR}, 2019, pp. 3585--3594.

\bibitem{nmp}
Y.~Hu, S.~Chen, X.~Chen, Y.~Zhang, and X.~Gu, ``Neural message passing for
  visual relationship detection,'' in \emph{ICMLW}, 2019.

\bibitem{girshick2015fast}
R.~Girshick, ``Fast r-cnn,'' in \emph{ICCV}, 2015, pp. 1440--1448.

\bibitem{he2017mask}
K.~He, G.~Gkioxari, P.~Doll{\'a}r, and R.~Girshick, ``Mask r-cnn,'' in
  \emph{ICCV}, 2017, pp. 2961--2969.

\bibitem{kim2016unified}
S.~Kim, K.~Park, K.~Sohn, and S.~Lin, ``Unified depth prediction and intrinsic
  image decomposition from a single image via joint convolutional neural
  fields,'' in \emph{ECCV}.\hskip 1em plus 0.5em minus 0.4em\relax Springer,
  2016, pp. 143--159.

\bibitem{misra2016cross}
I.~Misra, A.~Shrivastava, A.~Gupta, and M.~Hebert, ``Cross-stitch networks for
  multi-task learning,'' in \emph{CVPR}, 2016, pp. 3994--4003.

\bibitem{ladicky2014pulling}
L.~Ladicky, J.~Shi, and M.~Pollefeys, ``Pulling things out of perspective,'' in
  \emph{CVPR}, 2014, pp. 89--96.

\bibitem{wang2015towards}
P.~Wang, X.~Shen, Z.~Lin, S.~Cohen, B.~Price, and A.~L. Yuille, ``Towards
  unified depth and semantic prediction from a single image,'' in \emph{CVPR},
  2015, pp. 2800--2809.

\bibitem{kendall2018multi}
A.~Kendall, Y.~Gal, and R.~Cipolla, ``Multi-task learning using uncertainty to
  weigh losses for scene geometry and semantics,'' in \emph{CVPR}, 2018, pp.
  7482--7491.

\bibitem{zhang2018joint}
Z.~Zhang, Z.~Cui, C.~Xu, Z.~Jie, X.~Li, and J.~Yang, ``Joint task-recursive
  learning for semantic segmentation and depth estimation,'' in \emph{ECCV},
  2018, pp. 235--251.

\bibitem{cai2018cascade}
Z.~Cai and N.~Vasconcelos, ``Cascade r-cnn: Delving into high quality object
  detection,'' in \emph{CVPR}, 2018, pp. 6154--6162.

\bibitem{shao2019objects365}
S.~Shao, Z.~Li, T.~Zhang, C.~Peng, G.~Yu, X.~Zhang, J.~Li, and J.~Sun,
  ``Objects365: A large-scale, high-quality dataset for object detection,'' in
  \emph{ICCV}, 2019, pp. 8430--8439.

\bibitem{lin2014microsoft}
T.-Y. Lin, M.~Maire, S.~Belongie, J.~Hays, P.~Perona, D.~Ramanan,
  P.~Doll{\'a}r, and C.~L. Zitnick, ``Microsoft coco: Common objects in
  context,'' in \emph{ECCV}.\hskip 1em plus 0.5em minus 0.4em\relax Springer,
  2014, pp. 740--755.

\bibitem{uav_github}
T.~Li, J.~Liu, W.~Zhang, Y.~Ni, W.~Wang, and Z.~Li, ``Uav-human: A large
  benchmark for human behavior understanding with unmanned aerial vehicles.''
  \url{https://github.com/SUTDCV/UAV-Human}, 2021.

\bibitem{baradel2018glimpse}
F.~Baradel, C.~Wolf, J.~Mille, and G.~W. Taylor, ``Glimpse clouds: Human
  activity recognition from unstructured feature points,'' in \emph{CVPR},
  2018, pp. 469--478.

\bibitem{ppdet2019}
P.~Authors, ``Paddledetection, object detection and instance segmentation
  toolkit based on paddlepaddle.''
  \url{https://github.com/PaddlePaddle/PaddleDetection}, 2019.

\bibitem{liu2017skeleton}
J.~Liu, A.~Shahroudy, D.~Xu, A.~C. Kot, and G.~Wang, ``Skeleton-based action
  recognition using spatio-temporal lstm network with trust gates,''
  \emph{TPAMI}, vol.~40, no.~12, pp. 3007--3021, 2017.

\bibitem{gao2019optimized}
X.~Gao, W.~Hu, J.~Tang, J.~Liu, and Z.~Guo, ``Optimized skeleton-based action
  recognition via sparsified graph regression,'' in \emph{ACM Multimedia},
  2019, pp. 601--610.

\bibitem{liu2017skeletonbase}
J.~Liu, G.~Wang, L.-Y. Duan, K.~Abdiyeva, and A.~C. Kot, ``Skeleton-based human
  action recognition with global context-aware attention lstm networks,''
  \emph{TIP}, vol.~27, no.~4, pp. 1586--1599, 2017.

\bibitem{si2018skeleton}
C.~Si, Y.~Jing, W.~Wang, L.~Wang, and T.~Tan, ``Skeleton-based action
  recognition with spatial reasoning and temporal stack learning,'' in
  \emph{ECCV}, 2018, pp. 103--118.

\bibitem{veeriah2015differential}
V.~Veeriah, N.~Zhuang, and G.-J. Qi, ``Differential recurrent neural networks
  for action recognition,'' in \emph{CVPR}, 2015, pp. 4041--4049.

\bibitem{liu2017global}
J.~Liu, G.~Wang, P.~Hu, L.-Y. Duan, and A.~C. Kot, ``Global context-aware
  attention lstm networks for 3d action recognition,'' in \emph{CVPR}, 2017,
  pp. 1647--1656.

\bibitem{zhang2018adding}
P.~Zhang, J.~Xue, C.~Lan, W.~Zeng, Z.~Gao, and N.~Zheng, ``Adding attentiveness
  to the neurons in recurrent neural networks,'' in \emph{ECCV}, 2018, pp.
  135--151.

\bibitem{tang2018deep}
Y.~Tang, Y.~Tian, J.~Lu, P.~Li, and J.~Zhou, ``Deep progressive reinforcement
  learning for skeleton-based action recognition,'' in \emph{CVPR}, 2018, pp.
  5323--5332.

\bibitem{li2018co}
C.~Li, Q.~Zhong, D.~Xie, and S.~Pu, ``Co-occurrence feature learning from
  skeleton data for action recognition and detection with hierarchical
  aggregation,'' \emph{arXiv preprint arXiv:1804.06055}, 2018.

\bibitem{si2019attention}
C.~Si, W.~Chen, W.~Wang, L.~Wang, and T.~Tan, ``An attention enhanced graph
  convolutional lstm network for skeleton-based action recognition,'' in
  \emph{CVPR}, 2019, pp. 1227--1236.

\bibitem{zhang2019view}
P.~Zhang, C.~Lan, J.~Xing, W.~Zeng, J.~Xue, and N.~Zheng, ``View adaptive
  neural networks for high performance skeleton-based human action
  recognition,'' \emph{TPAMI}, vol.~41, no.~8, pp. 1963--1978, 2019.

\bibitem{baradel2017human}
{Baradel, Fabien and Wolf, Christian and Mille, Julien}, ``Human action
  recognition: Pose-based attention draws focus to hands,'' in \emph{ICCVW},
  2017, pp. 604--613.

\bibitem{baradel2018human1}
F.~Baradel, C.~Wolf, and J.~Mille, ``Human activity recognition with
  pose-driven attention to rgb,'' in \emph{BMVC}, 2018.

\bibitem{song2018skeleton}
S.~Song, C.~Lan, J.~Xing, W.~Zeng, and J.~Liu, ``Skeleton-indexed deep
  multi-modal feature learning for high performance human action recognition,''
  in \emph{ICME}.\hskip 1em plus 0.5em minus 0.4em\relax IEEE, 2018, pp. 1--6.

\bibitem{das2019toyota}
S.~Das, R.~Dai, M.~Koperski, L.~Minciullo, L.~Garattoni, F.~Bremond, and
  G.~Francesca, ``Toyota smarthome: Real-world activities of daily living,'' in
  \emph{ICCV}, 2019, pp. 833--842.

\bibitem{wheretofocus}
S.~{Das}, A.~{Chaudhary}, F.~{Bremond}, and M.~{Thonnat}, ``Where to focus on
  for human action recognition?'' in \emph{WACV}, 2019, pp. 71--80.

\bibitem{fan2020context}
Y.~Fan, S.~Weng, Y.~Zhang, B.~Shi, and Y.~Zhang, ``Context-aware
  cross-attention for skeleton-based human action recognition,'' \emph{IEEE
  Access}, vol.~8, pp. 15\,280--15\,290, 2020.

\bibitem{das2020vpn}
S.~Das, S.~Sharma, R.~Dai, F.~Bremond, and M.~Thonnat, ``Vpn: Learning
  video-pose embedding for activities of daily living,'' in \emph{ECCV}.\hskip
  1em plus 0.5em minus 0.4em\relax Springer, 2020, pp. 72--90.

\bibitem{wang2013learning}
J.~Wang, Z.~Liu, Y.~Wu, and J.~Yuan, ``Learning actionlet ensemble for 3d human
  action recognition,'' \emph{TPAMI}, vol.~36, no.~5, pp. 914--927, 2013.

\bibitem{lee2017ensemble}
I.~Lee, D.~Kim, S.~Kang, and S.~Lee, ``Ensemble deep learning for
  skeleton-based action recognition using temporal sliding lstm networks,'' in
  \emph{ICCV}, 2017, pp. 1012--1020.

\bibitem{rahmani2015learning}
H.~Rahmani and A.~Mian, ``Learning a non-linear knowledge transfer model for
  cross-view action recognition,'' in \emph{CVPR}, 2015, pp. 2458--2466.

\bibitem{rahmani20163d}
{Rahmani, Hossein and Mian, Ajmal}, ``3d action recognition from novel
  viewpoints,'' in \emph{CVPR}, 2016, pp. 1506--1515.

\bibitem{li2020hard}
T.~Li, J.~Liu, W.~Zhang, and L.~Duan, ``Hard-net: Hardness-aware discrimination
  network for 3d early activity prediction,'' in \emph{ECCV}.\hskip 1em plus
  0.5em minus 0.4em\relax Springer, 2020, pp. 420--436.

\end{thebibliography}

\vfill

\end{document}